\documentclass[conference]{IEEEtran}

\usepackage{subfigure}
\ifCLASSINFOpdf
   \usepackage[pdftex]{graphicx}
  \graphicspath{{../pdf/}{../jpeg/}}
   \DeclareGraphicsExtensions{.pdf,.jpeg,.png}
\else
\fi

\usepackage{times}
\usepackage{epsfig}
\usepackage{graphicx}
\usepackage{amsmath}
\usepackage{amssymb}
\usepackage{subfig}
\usepackage{alphalph}
\usepackage{algorithm}
\usepackage[noend]{algpseudocode}
\usepackage{tabularx}

\makeatletter
\def\BState{\State\hskip-\ALG@thistlm}
\makeatother

\begin{document}
%
\title{Multi-Scale Saliency Detection using Dictionary Learning}

\author{\IEEEauthorblockN{Shubham Pachori}
\IEEEauthorblockA{Electrical Engineering\\
Indian Institute of Technology \\
Gandhinagar, India 382355\\
Email: shubham\textunderscore pachori@iitgn.ac.in}}


%


\maketitle

\begin{abstract}
Saliency detection has drawn a lot of attention of researchers in various fields over the past several years. Saliency is the perceptual quality that makes an object, person to draw the attention of humans at the very sight. Salient object detection in an image has been used centrally in many computational photography and computer vision applications like video compression \cite{itti2004automatic}, object recognition and classification \cite{navalpakkam2006integrated}, object segmentation \cite{liu2012unsupervised}\cite{han2006unsupervised}\cite{ko2006object}, adaptive content delivery \cite{ma2003contrast}, motion detection \cite{liu2009object}, content aware resizing \cite{avidan2007seam}, camouflage images \cite{chu2010camouflage} and change blindness images \cite{ma2013change} to name a few. We propose a method to detect saliency in the objects using multimodal dictionary learning which has been recently used in classification \cite{bahrampour2016multimodal} and image fusion \cite{kim2016joint}. The multimodal dictionary that we are learning is task driven which gives improved performance over its counterpart (the one which is not task specific). 
\end{abstract}


%
\IEEEpeerreviewmaketitle

\section{Introduction}
We propose to detect salient regions in an image using multimodal dictionary regression method. We propose to detect the salient regions at different scales of an image because we believe that salient region should be able to gather attention of the viewer at different scales also. Moreover, it is also important to exploit the statistics of an image, cues which determine a region to be salient like texture, RGB, L*a*b, HSV values of the image and local binary features of the image, which could distinguish whether the contrast of a region is higher than that of the background thus deciding whether to mark a region as salient or not. These cues, when integrated at different scales of an image could be very helpful in deciding which regions of the image are salient. For the final step, we propose to use the regression by using the multimodal dictionary learning method. Here, different modes refer to the saliency map at different scale levels of the image. As detecting saliency map at each image scale is independent of the other, except for forming the super pixels for the multi-level segmentation hence these constituent different modes. Because dictionary learning is a patch-wise method, thus, it could take into the account the neighboring pixels while fusing the saliency maps at different scales, unlike regression algorithm which generally performs on pixel-to-pixel basis. Dictionary methods exploits sparse matrix for non-linearity, thus it is faster both in computations and training than several other non-linear machine learning algorithms like artificial neural networks, decision trees and support vector machines. 
\section{Related Work}

The saliency detection techniques could be broadly classified into two categories \cite{zhang2010adaptive}: bottom-up and top-down saliency detection. In bottom-up models, low-level visual features such as color, texture, orientation and intensity etc. are extracted from the image and are combined into saliency map. Salient locations are then identified using inhibition of return and winner-take-all operations \cite{itti1998model}. Top-down models are task-dependent and use a priori knowledge of the visual system. They are integrated with the bottom-up models to generate saliency maps for localizing the objects of interest. Among bottom-up approaches, in \cite{itti1998model}, a biologically plausible saliency detection method based on color contrast called "center - surround difference" is proposed. A graph-based visual saliency (GBVS) model based on the Markovian approach is suggested in \cite{harel2006graph}. In \cite{yan2013hierarchical}, contrast features of different scales of an image were computed using hierarchical model and fused togeher into a single map using a graphical model. The negative logarithm of the probability known as Shannnon's self-information is used to measure the improbability of a local patch as a bottom-up saliency cue in \cite{zhang2008sun} and \cite{bruce2009saliency}. In \cite{liu2012unsupervised}, a kernel estimation (KDE) based non parametric model was constructed for each segmented region, and color and spatial saliency measures of KDE models were evaluated and used to measure saliency of pixels. \\
Some models also exploit machine learning techniques to learn saliency. In \cite{judd2009learning} authors used a combination of low, middle and high level features, and the saliency classification is done in pixel-by-pixel by manner by training a Support Vector Machines. In \cite{zhao2015saliency}, deep features are exploited to learn multi-context saliency models. In \cite{jiang2013salient}, random forests are used to calculate the saliency of a region in image. Another kind of saliency models are based on spectral analysis [\cite{hou2007saliency},\cite{li2013visual},\cite{goferman2012context},\cite{hou2012image}]. Some models also leverage the topological structure of a scene for saliency detection. In \cite{wei2012geodesic}, a local patch's saliency is measured on a graphical model by its shortest distance to the image borders. The weights of the edges in a graphical model are computed based on local dissimilarity. In \cite{zhang2010adaptive}, the attention of Gaussian Mixture Models (GMM)for salient object and background GMM were constructed based on the image clustering result, and pixels were classified under the Bayesian framework to obtain the salient object. In \cite{gopalakrishnan2009salient}, a framework of mixture of Gaussian in H-S space was used to compute the distance between clusters and color spatial distribution which were then utilized to selectively generate a saliency map. In \cite{xie2013bayesian}, a bayesian framework was proposed to combine mid-level cues (saliency information provided by superpixels) and the low-level cues (coarse saliency region obtained via convex hull of interest points) and to generate a saliency map.\\
There have been works in the context aware and statistical-based models for saliency too. In \cite{wang2011automatic}, a trained classifier called the auto-context model is used to enhance an appearance - based optimization framework for salient region framework for salient region detection. In \cite{itti2004automatic}, patch - based dictionary learning is used for rarity-based saliency model. In \cite{borji2015salient}, summary of more works on saliency detection could be found out.

\section{Paper Organization}

To make the reader feel comfortable  about the notations, we have used the similar notation and variables as mentioned in the papers \cite{bahrampour2016multimodal} and \cite{jiang2013salient}, so that they could check those papers in case he/she feels anything missing in our report or cannot understand my approach from this report alone. In the first half of the proposed approach, we have discussed that how we have obtained the saliency score maps which are then merged using multimodal dictionary methods.

\section{Notation for the Section V.B}

Vectors are denoted by bold case letters and matrices by upper case letters. For a given vector ${\textit{x}}$, $x_{i}$ is its $i^{th}$ element. For a given finite set of indices $\gamma$, ${x}_{\gamma}$ is the vector formed with those elements of ${x}$ indexed in $\gamma$. Symbol $\rightarrow$ is used to distinguish the row vectors from column vectors, i.e. for a given matrix {${\textit{X}}$}, the $i^{th}$ row and $j^{th}$ column of matrix are represented as ${\textit{x}}_{i\rightarrow}$ and ${\textit{x}}_{j}$, respectively. For a given finite set of indices $\gamma$, ${\textit{X}}_{\gamma}$ is the matrix formed with those columns of ${\textit{X}}$ indexed in $\gamma$ and ${\textit{X}}_{\gamma\rightarrow}$ is the matrix formed with those rows of ${\textit{X}}$ indexed in $\gamma$.
Similarly, for given finite sets of indices $\gamma$ and $\Psi$, ${\textit{X}}_{\gamma\rightarrow,\Psi}$ is the matrix formed with those rows and columns of ${\textit{X}}$ indexed in $\gamma$ and $\Psi$, respectively. $x_{ij}$ is the element of ${\textit{X}}$ at row $i$ and column $j$. The $l_{q}$ norm, $q \geq 1$, of a vector ${\textit{x}}$ $\epsilon$ $R^{m}$ is defined as $||{\textit{x}}||_{l_{q}} = (\sum^{m}_{j = 1}|x_{j}|^{q})^{\frac{1}{q}}$. The Frobenius norm and $l_{1q}$ norm, $q \geq 1$, of matrix ${\textit{X}}$ $\epsilon$ $R^{m\times n}$ is defined as $||{\textit{X}}||_{F} = (\sum^{m}_{i = 1}\sum^{n}_{j = 1}x_{ij}^{2})^{\frac{1}{2}}$ and $||{\textit{X}}||_{l_{1q}} = \sum^{m}_{j = 1}||{\textit{x}}_{{i\rightarrow}}||_{l_{q}}$  respectively. The collection {$\lbrace{\textit{x}}^{i}|i \epsilon \gamma\rbrace$}  is shortly denoted as $\lbrace{\textit{x}}^{i}\rbrace$.

\section{Proposed Method}

The proposed is very much similar to the saliency detection method proposed in \cite{jiang2013salient} but differs in the way that we compute saliency at different scale-space and instead of using uniform weights for all pixels in an image for obtaining the final saliency map, we leverage dictionary learning method for combining the pixels in a non-linear fashion. Our method of detection salient regions in images is divided into three main parts. The first part consists of constructing the gaussian scale space of an image and clustering the similar regions of images at each gaussian scale level into super pixels. Then the likelihood of each superpixel being salient at each gaussian scale level is calculated using random forests. The final step consists of fusing the images at different scale space to obtain the final results using multimodal dictionary regression. The whole process has been depicted in Fig 1. 

\begin{figure*}
\label{pipeline}
\centering
\includegraphics[scale=0.34]{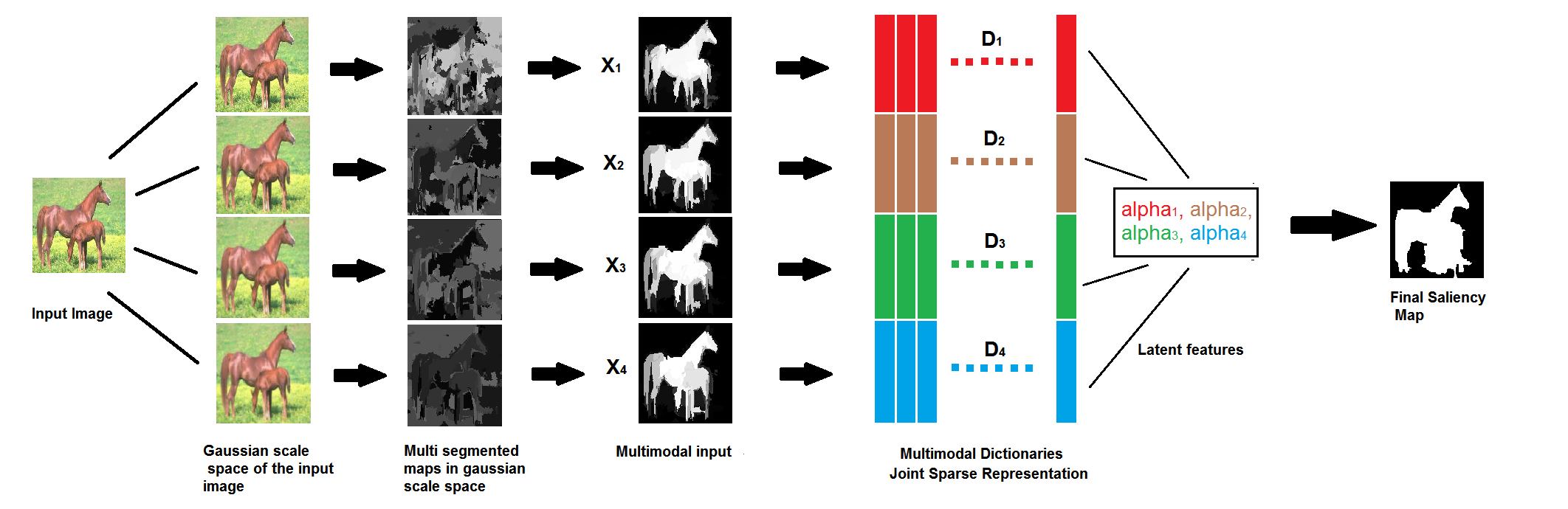}
\caption{The pipeline of the method that we have adopted.}
\end{figure*}

\subsection{Obtaining the saliency scores of regions of images at different scales}

\subsubsection{{Obtaining gaussian scale space of the image}}

We think that if a region is salient then it must also attract the eyesight of an observer at different scale space. The gaussian scale space of an image is equivalent to convolving an image successively with a gaussian filter \cite{adelson1984pyramid}. Rather than reducing the size at each level, we therefore directly convolve the image with a gaussian filter at each scale space. To obtain scale space we used a gaussian filter with blur radius $\sigma = 1.2$. Let the depth of the scale space be equal to $M$. For our experiments, we use $M = 4$.  

\subsubsection{{Clustering the pixels into super pixels at each level of gaussian pyramid}}

We follow the procedure as mentioned in \cite{jiang2013salient} to produce the multi-level segmentations $[S_{1}, S_{2} ..., S_{M}]$, where $M$ is the depth of the gaussian pyramid, and each segmentation $S_{i}$ corresponds to the segmentation at $i^{th}$ depth of the gaussian scale-space of the image $I$. Since, we have formed a gaussian scale space thus, $S_{1}$ and $S_{M}$ are the finest and coarsest segmentation respectively. Other segmentations [$S_{2},..., S_{M}$] are dependent upon $S_{1}$ as they are computed by merging the segmentation regions in each layer of the pyramid successively i.e. $S_{k}$ depends upon $S_{k-1}$. Regional pairs are merged sequentially as per the similarity of the corresponding regions only if the similarity of two regions is greater than the specified threshold. The procedure is shown in Fig. 2. More details could be found in \cite{felzenszwalb2004efficient}.

\begin{figure}
\label{segmentation}
\centering
\includegraphics[scale=0.42]{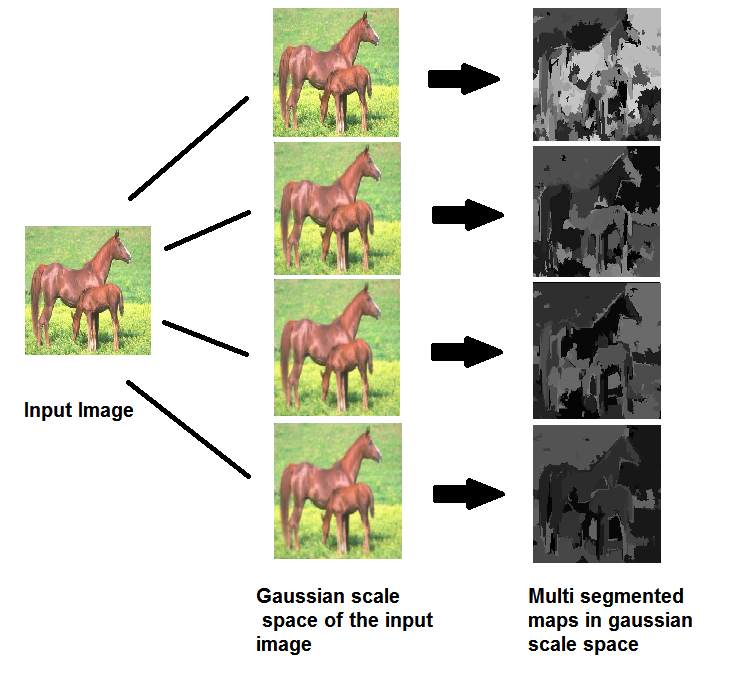}
\caption{Clustering the regions of the images at different gaussian scale space levels into superpixels.}
\end{figure}

\subsubsection{{Computing the saliency scores of the different regions}}

We use the cues mentioned in \cite{jiang2013salient} to compute the saliency scores of different regions. The authors computed the contrast descriptor of a region. The scores of the different regions are found at different scales to obtain the saliency maps at different scale as $D_{1},D_{2},...,D_{M}$. They extracted descriptors for a region based on its contrast difference between neighbourhood regions, its intrensic properties and its backgroundness. 
The tables given in Fig. 3 and Fig. 4 depict the descriptors that were extracted from a superpixel region. More information on how to extract the regional features could be found in \cite{jiang2013salient}.\\
Each region is described by a feature vector $\textbf{\textit{v}}$, composed of the regional contrast, regional property, and regional backgroundness descriptors. A pre-trained random forest regressor $\textbf{\textit{F}}$ \cite{jiang2013salient} is used which estimates the saliency score $\textbf{\textit{A}}$ of a region using the features $\textbf{\textit{v}}$ of a region. Learning a saliency regressor can automatically combine the features and discover the most discriminative ones. The process is given in Fig. 5. Once the saliency score of each region of each image at different gaussian scale space has been computed the next task is to merge them non-linearly and patch wise to produce the final output image.

\begin{figure*}
\label{figtae1}
\centering
\includegraphics[scale=0.75]{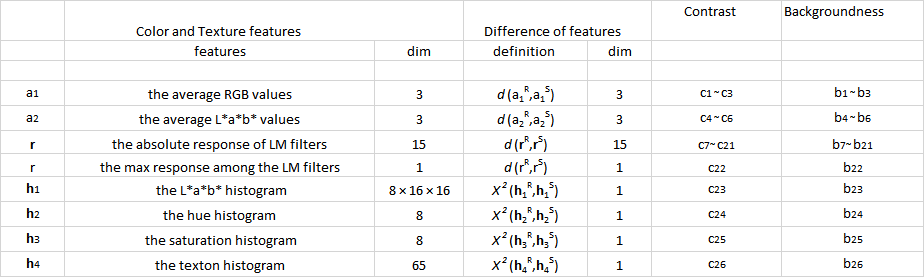}
\caption{Color and texture features. $d(x_{1}, x_{2}) = ( |x_{11} - x_{21}|,. . . , |x_{1k} - x_{2k}| )$, where $k$ is the number of elements in the vectors $x_{1}$ and $x_{2}$. $\textit{X}^{2}(h_{1},h_{2}) = \sum_{i=1}^{k} \frac{2(h_{1i} - h_{2i})^{2}}{h_{1i} + h_{2i}} $ with $k$ being the number of histogram bins The last two columns denote the symbols for regional contrast and backgroundness descriptors }
\end{figure*}

\begin{figure*}\label{figtable2}
\centering
\includegraphics[scale=0.685]{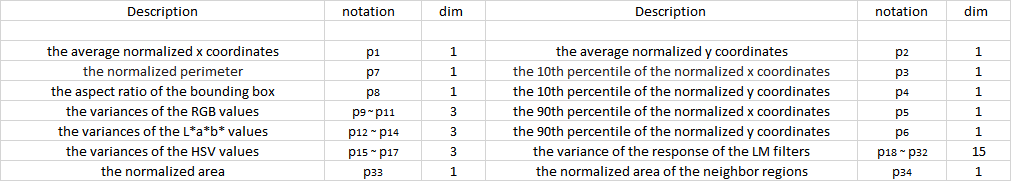}
\caption{The regional property descriptors.}
\end{figure*}

\begin{figure}
\label{figtae1}
\centering
\includegraphics[scale=0.26]{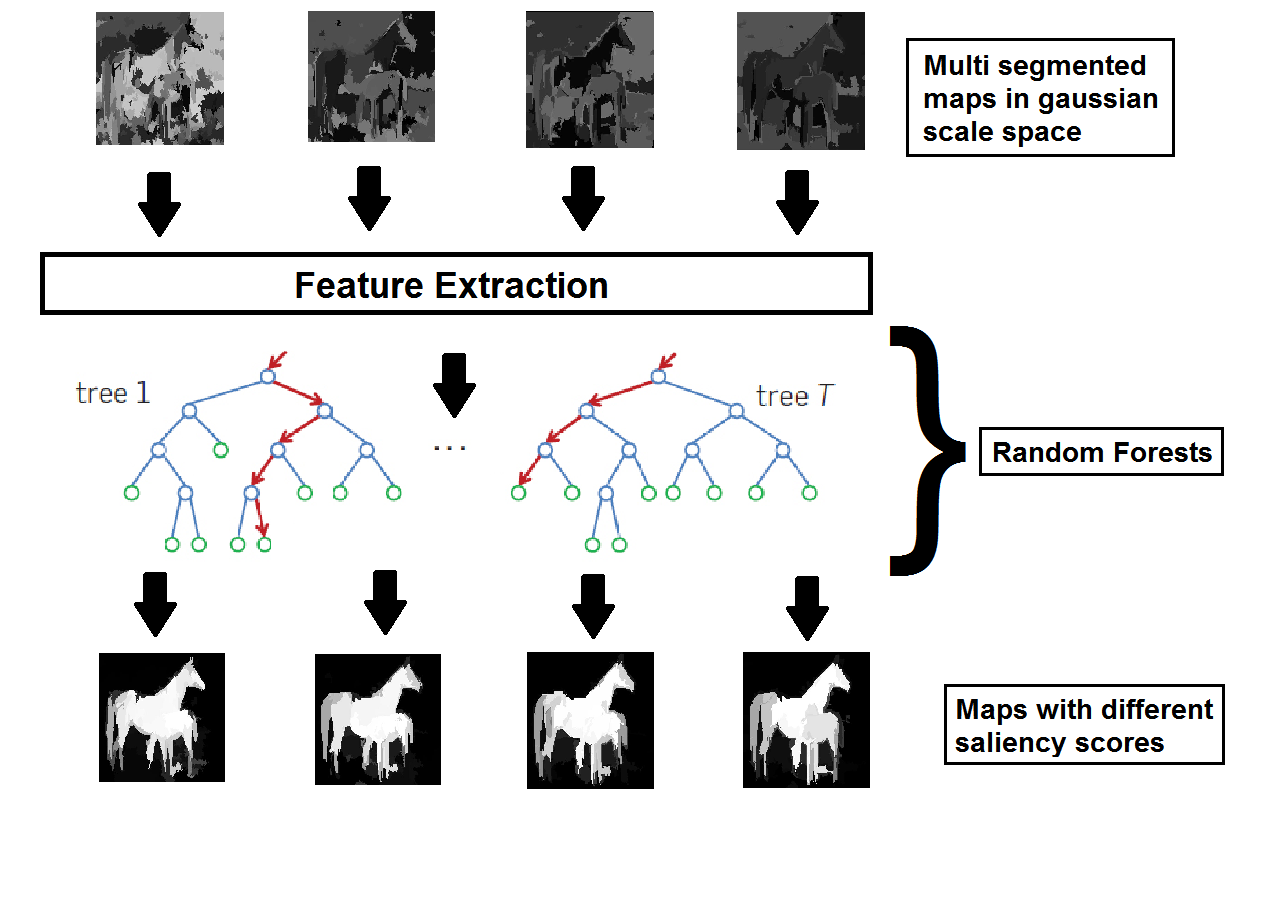}
\caption{Computing the saliency scores of different regions of multi-segmented images at different gaussian scale space using random forest regressor.}
\end{figure}

\subsection{Merging the saliency maps to obtain the final saliency map}

To obtain the final saliency map, we extract $(9 \times 9)$ size patch from each saliency map $D_{1},D_{2},...,D_{M}$ and merge the corresponding patches of $D_{1},D_{2},...,D_{M}$. We have extracted patches of size $(5 \times 5)$, $(7\times 7)$, $(9 \times 9)$, $(11 \times 11)$ and $(13\times 13)$ but found experimentally that the best results were obtained when we extracted the patches of size $(9 \times 9)$. The extracted patches were arranged in a lexicographical order to form a column vector of size $(81\times M)$. The column vector was then used to produce a column vector of size $(81 \times 1 )$, which is then reshaped to the size $(9 \times 9 )$  of the final saliency map. To obtain the final saliency map, we trained multimodal dictionary specifically to obtain the binary maps using the saliency obtained maps at different scale space.

\subsubsection{Training Multimodal Dictionary Learning}
The unsupervised method of dictionary learning which minimizes the reconstruction error, does not take into the account the output labels obtained through ground truth while training which leads to the poorer performance in the output. Here, we following the method of [34], train the dictionary in the supervised fashion. Here, a multimodal task-driven dictionary learning algorithm is proposed that enforces collaboration among the modalities both at the feature level using joint sparse representation and the decision level using a sum of the decision scores. We propose to learn the dictionaries $D$, $\forall s \in M$, and the classifier parameters  $\forall s \in M$, shortly denoted as the set ${D, w}$, jointly as the solution of the following optimization problems:
\begin{equation}
\label{eqn_example}
l_{un}({\textit{x, D}}) = \frac{1}{2} \sum^{M}_{s=1} || {\textit{x}} - { \textit{D} } \alpha_{s} ||^{2}_{l_{2}} + \lambda_{1}||{\textit{A}}||_{l_{12}} + \frac{\lambda_{2}}{2}||{\textit{A}}||^{2}_{F}
\end{equation}
\begin{equation}
l_{su}(y, W, \alpha^{*}) = \frac{1}{2}||y - W\alpha^{*}||^{2}_{l_{2}} + \frac{\nu}{2}||w||^{2}_{l_{2}}
\label{eqn_example}
\end{equation}
Here, the label \textit{{un}} and \textit{{su}} emphasizes the learning in unsupervised and supervised formulation respectively. $\lambda_{1}$, $\lambda_{2}$ are the regularization parameters for controlling sparsity and overfitting respectively. In unsupervised learning we obtain the matrix $A^{*}$ as [$\alpha^{*1} , \alpha^{*2} ... \alpha^{*M}$], which minimizes the Eq. 1 given a dictionary $D$ and weight matrix $w$. \\ 
Another use of a given trained dictionary is for feature extraction where the sparse code $A^{*}(x, D)$ obtained as a solution of Eq. 1 is used as a feature vector representing the input signal $x$ in the classical expected risk optimization for training the regressor given in the equation Eq. 2. In equation Eq. 2, $y$ is the ground truth class label corresponding to  the input $x$, $w$ is model (classifier) parameters, $\nu$ is a regularizing parameter, which controls the overfitting. $l$ is a convex loss function that measures the error of the obtained output $y$ given the feature matrix $A^{*}$, and model parameters $(D,w)$. Note that in Eq. 2, the dictionary $D$ is fixed and independent of the given task and class label $y$. In task-driven dictionary learning, on the other hand, a supervised formulation is used which finds the optimal dictionary and classifier parameters jointly by solving the following optimization problem \cite{mairal2012task}.
\begin{equation}
\label{eqn_example}
\min_{D \in D, w\in W} =  E_{y,x} [l_{su} (y, w, \alpha^{*}(x,D))] + \frac{\nu}{2} ||{w}||_{l_{2}}^{2}
\end{equation}
Our problem could be formalized as solving the following optimization problem.
\begin{equation}
\label{eqn_example}
\min_{D \in D, w\in W} =  f(\lbrace D^{s},w^{s}\rbrace) + \frac{\nu}{2} \sum^{M}_{s=1} ||{w}||_{l_{2}}^{2}
\end{equation}
where $f$ is defined as the expected cumulative cost:
\begin{equation}
\label{eqn_example}
f(\lbrace D^{s},w^{s}\rbrace) = E \sum^{M}_{s=1} l_{su}(y,w^{s},\alpha^{s*}),
\end{equation}

\begin{figure*}\label{fig:1}
\centering
\includegraphics[scale=0.45]{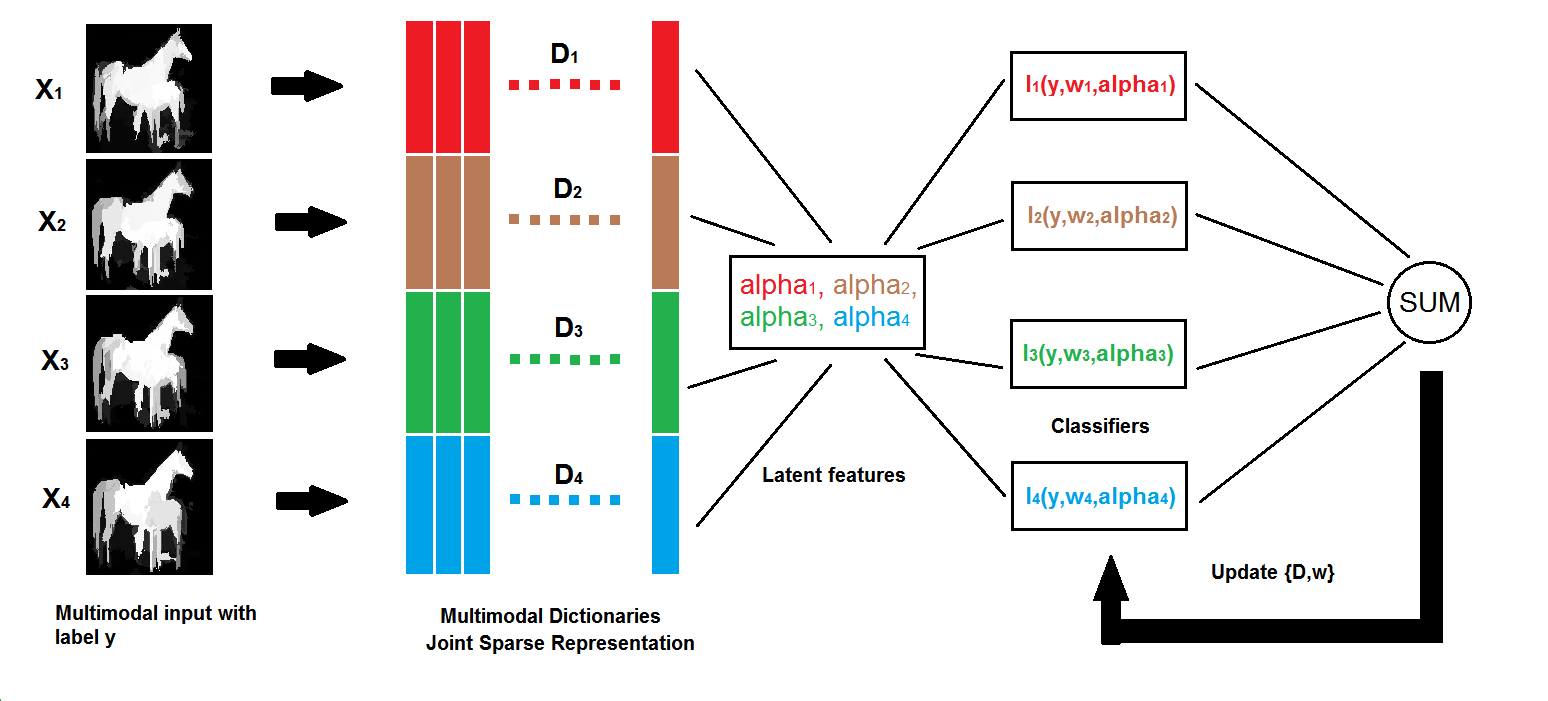}
\caption{Training multimodal dictionary and weights to fuse the maps at different gaussian scale space to obtain the final saliency map.}
\end{figure*}

Some assumptions are necessary to prove the differentiability of $f$, which are as follows:\\
\textbf{a)} The multimodal data $(y, \lbrace x^{s}\rbrace)$ admit a probability density $p$ with compact support.\\
\textbf{b)} For all possible values of $y, p(y, .)$ is continuous and $l_{su}(y, .)$ is twice continuously differentiable.\\
The steps involved to update the dictionary and weights go as below. For detailed information one could read the paper \cite{bahrampour2016multimodal}.\\
\textbf{Step 1:} Find the active set $\Lambda$ of the solution $A^{*}$ of the joint sparse coding problem as
\begin{equation}
\label{eqn_example}
\Lambda =  \lbrace j \in \lbrace 1,...,d \rbrace : \Vert a^{*}_{j\rightarrow} \Vert_{l_{2}} \neq 0 \rbrace,
\end{equation}
where $a^{*}_{j\rightarrow}$ is the $j^{th}$ row of $A^{*}$. \\
\textbf{Step 2:} Let $\lambda_{2} > 0$ and let $\Upsilon =  \cup_{j\in \Lambda} \Upsilon_{j}$ where $\Upsilon_{j} = \lbrace j, j+d, ..., j+(M-1)d \rbrace.$ Let the matrix $\hat{D} \in R^{n\times |\Upsilon|}$ be defined as 
\begin{equation}
\label{eqn_example}
\hat{D} = [ \hat{D}_{1}, \hat{D}_{2}, ..., \hat{D}_{|\Lambda|} ],
\end{equation}
where $\hat{D}_{j} = blkdiag(d^{1}_{j}, d^{2}_{j},..., d^{M}_{j}) \in R^{n \times M}, \forall j \in \Lambda,$ is the collection of the $j^{th}$ active atoms of the multimodal dictionaries, $d^{s}_{j}$ is the $j^{th}$ active atom of $D^{s}$, $blkdiag$ is the block diagonalization operator, and $n = \sum_{s\in M} n^{s}$.\\
\textbf{Step 3:} Let matrix $\Delta \in R^{|\Upsilon| \times |\Upsilon|}$ be defined as
\begin{equation}
\label{eqn_example}
\Delta = blkdiag( \Delta_{1}, \Delta_{2}, ..., \Delta_{|\Lambda|}),
\end{equation}
where \( \displaystyle \Delta_{j} = \frac{1}{||a^{*}_{j\rightarrow} ||_{l_{2}}} I - \frac{1}{||a^{*}_{j\rightarrow} ||_{l_{2}}^{3}} (a^{*}_{j\rightarrow})^{T} (a^{*}_{j\rightarrow}) \in R^{M\times M} \), $j \in \Lambda$ , and $I$ is the identity matrix. Then, the function $f$ defined in Eq. 5 is differentiable and $\forall s \in M,$
\begin{equation}
\label{eqn_example}
\nabla_{w^{s}} f = E[\nabla_{w^{s}} l_{su} (y,w^{s}, \alpha^{s*})]
\end{equation}
\begin{equation}
\label{eqn_example}
\nabla_{D^{s}} f = E[(x^{s} - D^{s}\alpha^{s*})\beta^{T}_{\bar{s}} - D^{s}\beta_{\bar{s}}\alpha^{s*^{T}})]  
\end{equation}
where $\bar{s} = \lbrace s, s+M, ..., s +(d-1)M  \rbrace$ and $\beta \in R^{dS}$ is defined as 
\begin{equation}
\label{eqn_example}
\beta_{\Upsilon^{c}} = 0, \hspace{0.1cm} \beta_{\Upsilon} = ({\hat{D}}^{T}{\hat{D}} + \lambda_{1} \Delta + \lambda_{2} I)^{-1}g  
\end{equation}
in which \( \displaystyle g = vec(\nabla_{A^{*}_{\Lambda \rightarrow}} \sum_{s=1}^{S} l_{su}(y_{t}, w^{s}, \alpha^{s*})) \), $\Upsilon^{c} = \lbrace 1, 2,..., dM \rbrace\setminus\Upsilon$, $\beta_{\Upsilon} \in R^{|\Upsilon|}$ is formed of those rows of $\beta$ indexed by $\Upsilon$, and $vec(.)$ is the vectorization operator.\\
The proof of the above steps could be found in detail in \cite{bahrampour2016multimodal}. The whole algorithm is summarized in Algorithm 1.

\begin{algorithm}
\caption{Multimodal Dictionary Learning}\label{euclid}
\begin{algorithmic}[1]
\State \textbf{Input:} Regularization parameters $\lambda_{1}$, $\lambda_{2}$, $\nu$, learning rate parameters $\rho, t_{0}$, number of iterations $T$, initial dictionaries $D_{s}$, initial weights and bias (model parameters) as $w_{s} \in W_{s}$
\State \textbf{Output:} Learned dictionary $D_{s}$ and model parameters $w_{s}$ 
\For{ $t = 1,...,T$ }
\State Draw a random sample $\lbrace x^{1}_{t} , x^{2}_{t} ... x^{S}_{t}, y_{t}\rbrace$ from the training data.
\State Find a solution $A^{*} = [\alpha^{*1},  \alpha^{*2}  ...  \alpha^{*S}] \in R^{d\times S} $ of the joint sparse coding problem.
\begin{equation}
\label{eqn_example}
l_{u}({\textit{x,D}}) = \frac{1}{2} \sum^{S}_{s=1} || {\textit{x}} - { \textit{D} } \alpha_{s} ||^{2}_{l_{2}} + \lambda_{1}||{\textit{A}}||_{l_{12}} + \frac{\lambda_{2}}{2}||{\textit{A}}||^{2}_{F}
\end{equation}
\State Compute set of active rows $\Lambda$ of $A^{*}$ as:
\begin{equation}
\label{eqn_example}
\Lambda =  \lbrace j \in \lbrace 1,...,d \rbrace : \Vert a^{*}_{j\rightarrow} \Vert_{l_{2}} \neq 0 \rbrace,
\end{equation}
\hspace{0.5cm} where $a^{*}_{j\rightarrow}$ is the $j^{th}$ row of $A^{*}$
\State Compute $\hat{D} \in R^{n\times|\Upsilon|}$ as given in Equation ().
\State Compute $\Delta \in R^{|\Upsilon|\times|\Upsilon|}$ as given in Equation ()
\State Compute $\beta \in R^{dS} as:$
\begin{equation}
\label{eqn_example}
\beta_{\Upsilon^{c}} = 0, \beta_{\Upsilon} = ({\hat{D}}^{T}{\hat{D}} + \lambda_{1} \Delta + \lambda_{2} I)^{-1}g 
\end{equation}
where $\Upsilon$ = $\bigcup_{j\in \Lambda} {j,j+d, ..., j+(S-1)d}$ and $g = vec(\nabla_{A^{*}_{\Lambda \rightarrow}} \sum_{s=1}^{S} l_{su}(y_{t}, w^{s}, \alpha^{s*})) $
\State Choosing the learning rate $\rho_{t} \leftarrow min(\rho, \rho\frac{t_{0}}{t})$
\State Update the parameters by a projected gradient step:
\begin{equation}
\label{eqn_example}
w^{s} \leftarrow \prod [w^{s} - \rho_{t}(\nabla_{w^{s}} l_{su}(y_{t}, w^{s}, \alpha^{s*}) + \nu w^{s})]
\end{equation}
\begin{equation}
\label{eqn_example}
D^{s} \leftarrow \prod [D^{s} - \rho_{t}(( x_{t}^{s} - D^{s}\alpha^{s*})\beta^{T}_{\bar{s}} - D^{s} \beta_{\bar{s}}(\alpha^{s*})^{T})]
\end{equation}
$\forall s \in S, where \bar{s} = {s, s+S, s+2S,...., (s+(M-1)S)} $
\EndFor
\end{algorithmic}
\end{algorithm}

\section{Experimental Results}

In this section we discuss the results obtained through our method for detecting the saliency. The final results are dependent on the hyperparameters: chosen. We chose only the benchmark dataset due to the lack of time (Though the results on other datasets too will be compared later as a future work). The size of dictionary $D$ that is taken for the experiments is $324 \times 150$, whereas the size of the weight matrix is take as $81 \times 600$ and size of the bias vector is $81 \times 1$. To obtain the value of $\alpha $ in Eq. 1. we used the SPAMS toolbox \cite{mairal2014sparse}. The hyperparameters in Eq. 1 are chosen as follows: $\lambda_{1} = 0.015$ and $\lambda_{1} = 0.002$. Th dictionary and weights are trained using the training vector of size $324 \times 181476.$ To obtain the labels we thresholded the ground. The pixel intensity was set to 1 if it was greater than 0.3 else was set to zero. We created a label matrix consisting of only 0's and 1's of size $81 \times 181476$.

\textbf{ECCSD}: This dataset consists of 1000 images containing the pixel wise ground truth annotation of salient objects. The dataset contains images each obtained in variety of scenarios and each containing variety of objects. This particular dataset was chosen for the evaluation of the method in the project because unlike other datasets it does not contain objects which are at he same time likely to be salient as marked as ground truth. In short there is no ambiguity between the ground truth of the salient regions in the images.

To evaluate the performance we used the precision-recall (PR) measure, Receiver-operator characteristics (ROC) measure and F-measure (the weighted harmonic mean of precision and recall) \cite{li2013contextual}. The formula for each is given as follows :
\begin{equation}
Precision = \frac{t_{p}}{t_{p} + f_{p}}
\end{equation}
\begin{equation}
Recall = \frac{t_{p}}{t_{p} + f_{n}}
\end{equation}
\begin{equation}
F-measure = \frac{2*Precision*Recall}{(Precision+Recall)}
\end{equation}

To obtain the data points for the curve, the saliency threshold that determines if a pixel is varied. The curve for each measure has been plotted below. We compared the results with BMS \cite{zhang2013saliency} , CA \cite{goferman2012context}, CB \cite{jiang2011automatic}, chm \cite{li2013contextual}, COV \cite{erdem2013visual}, DRFI \cite{jiang2013salient}, DSR \cite{li2013saliency}, FES \cite{tavakoli2011fast}, FT \cite{achanta2009frequency}, GB \cite{harel2006graph}, GC \cite{cheng2013efficient}, GMR \cite{yang2013saliency}, GU, HDCT \cite{kim2014salient}, IT \cite{itti1998model}, LMLC \cite{xie2013bayesian}, MC \cite{jiang2013saliency}, MNP \cite{margolin2013saliency}, MSS \cite{achanta2010saliency}, PCA \cite{margolin2013makes}, QCUT \cite{aytekin2014automatic} , rbd \cite{zhu2014saliency}, SEG \cite{rahtu2010segmenting}, SeR \cite{seo2009static}, SF \cite{perazzi2012saliency}, SIM \cite{murray2011saliency}, SUN \cite{zhang2008sun}, SWD \cite{duan2011visual}, ufo \cite{jiang2013salient}. The saliency maps for these are obtained from the project page of the paper \cite{borji2015salient}.

\begin{figure}\label{PR Curve}
\centering
\includegraphics[scale=0.2]{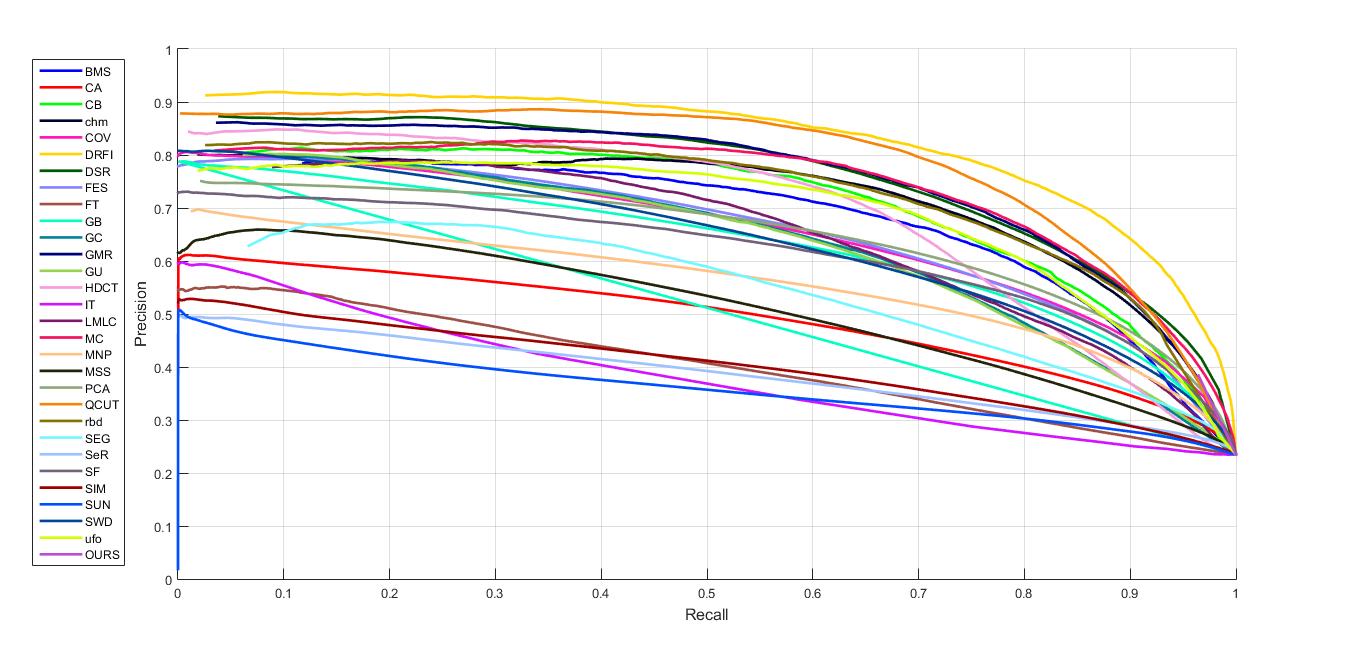}
\caption{Comparison of Precision-Recall between various methods proposed for saliency detection.}
\end{figure}

\begin{figure}\label{Fmeasure}
\centering
\includegraphics[scale=0.2]{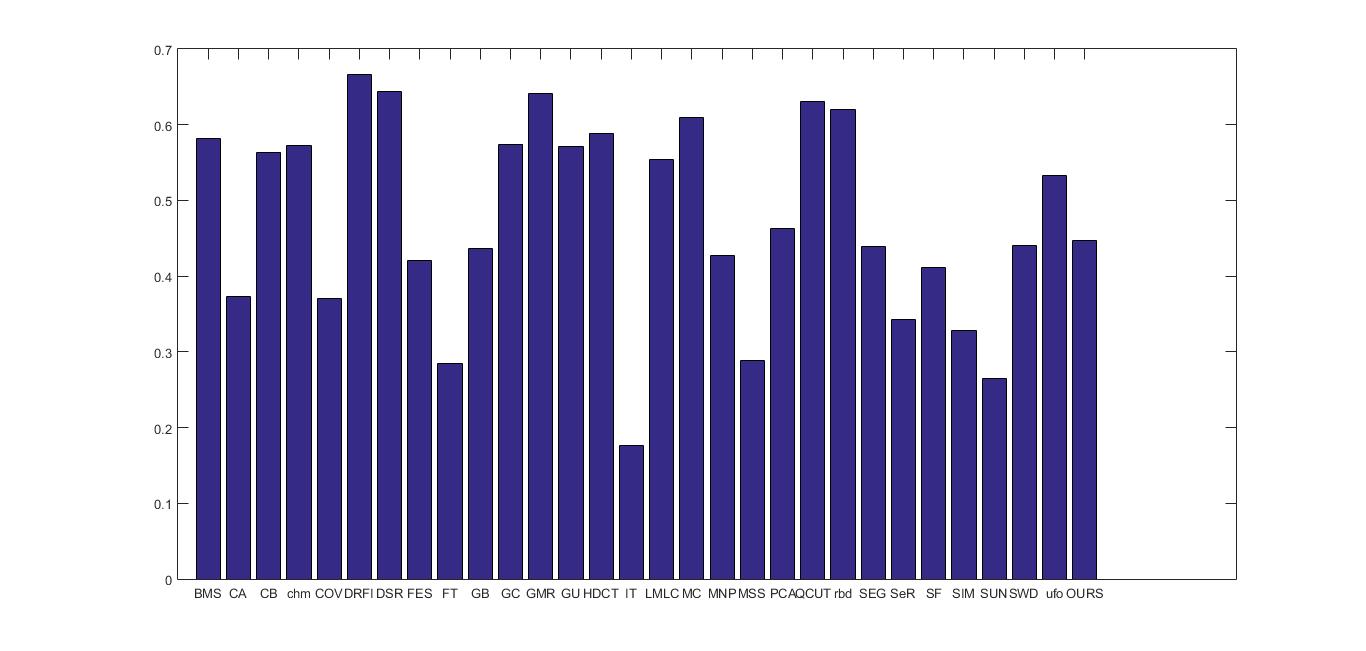}
\caption{Comparison of Fmeasure between various methods proposed for saliency detection.}
\end{figure}

\begin{figure}\label{ROCurve}
\centering
\includegraphics[scale=0.2]{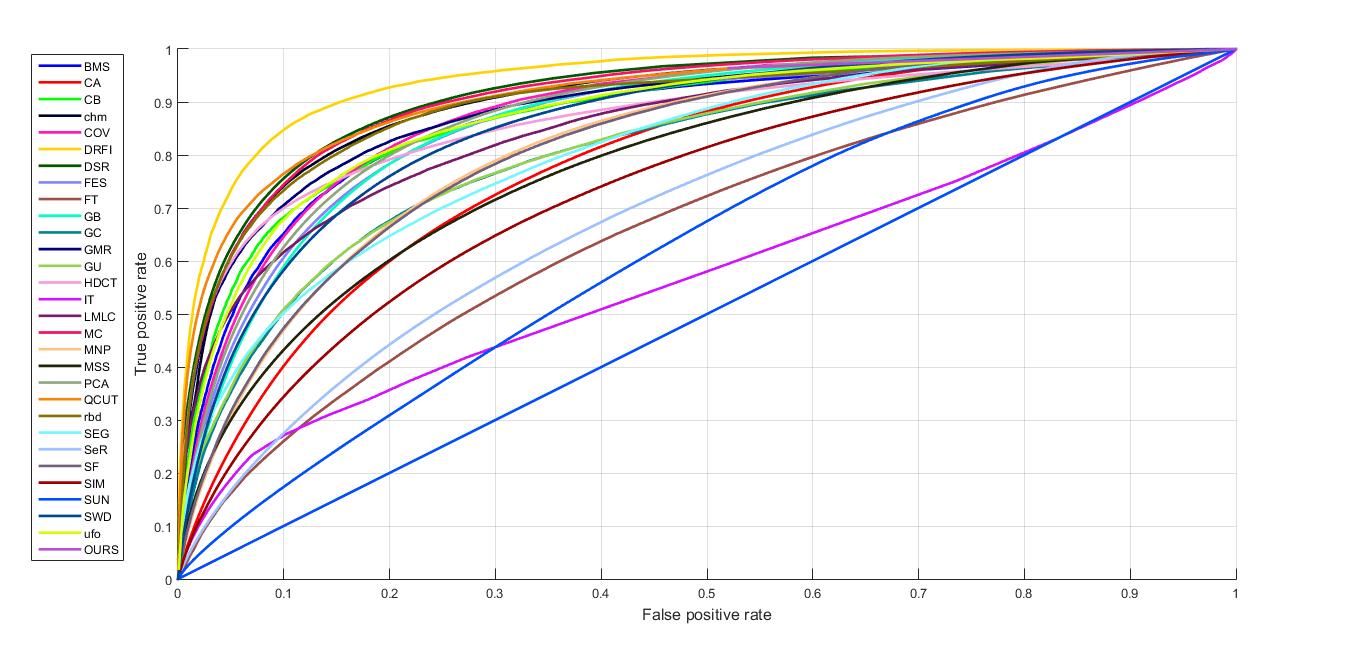}
\caption{Comparison of ROC between various methods proposed for saliency detection.}
\end{figure}

\begin{figure*}
\centering
\subfigure{\includegraphics[width=0.1\textwidth]{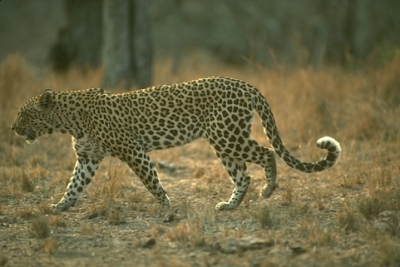}}%
\subfigure{\includegraphics[width=0.1\textwidth]{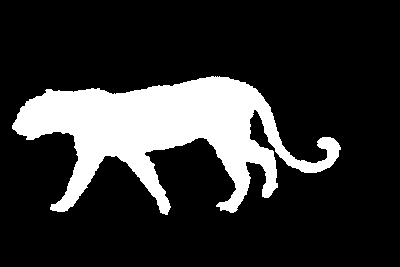}}%
\subfigure{\includegraphics[width=0.1\textwidth]{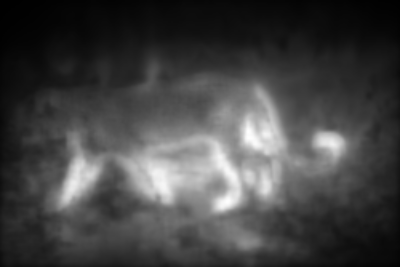}}%
\subfigure{\includegraphics[width=0.1\textwidth]{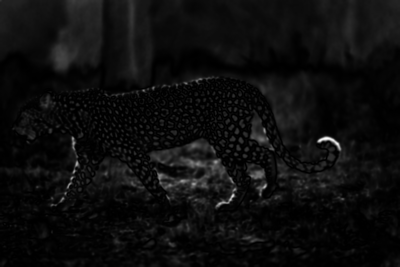}}%
\subfigure{\includegraphics[width=0.1\textwidth]{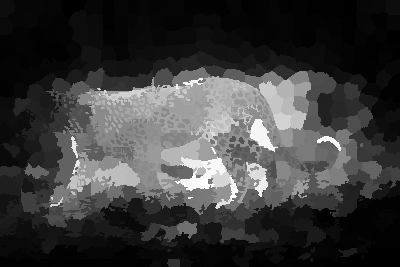}}%
\subfigure{\includegraphics[width=0.1\textwidth]{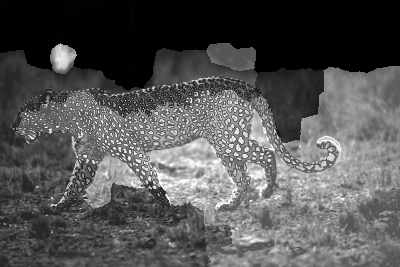}}%
\subfigure{\includegraphics[width=0.1\textwidth]{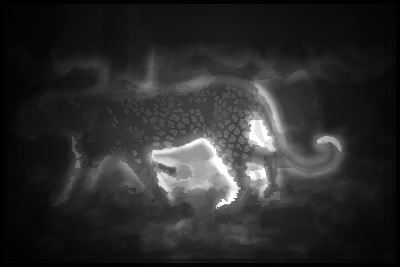}}%
\subfigure{\includegraphics[width=0.1\textwidth]{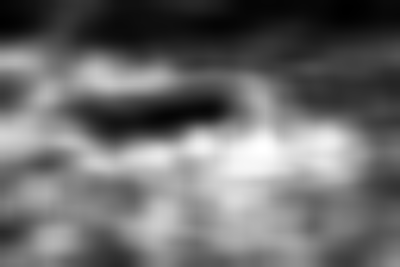}}%
\subfigure{\includegraphics[width=0.1\textwidth]{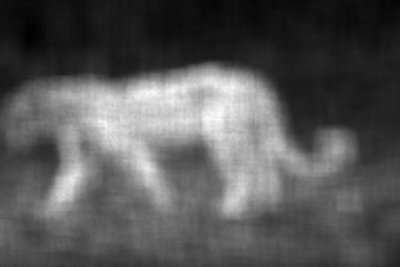}}%
\subfigure{\includegraphics[width=0.1\textwidth]{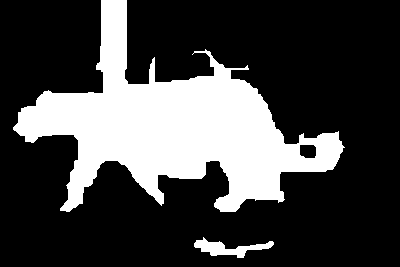}}\\
\subfigure{\includegraphics[width=0.1\textwidth]{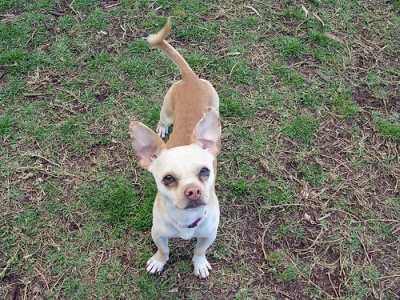}}%
\subfigure{\includegraphics[width=0.1\textwidth]{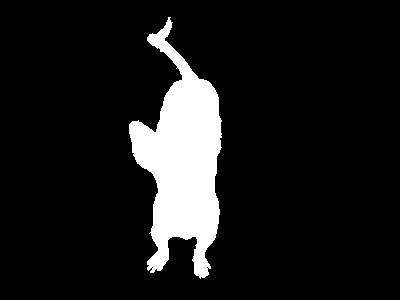}}%
\subfigure{\includegraphics[width=0.1\textwidth]{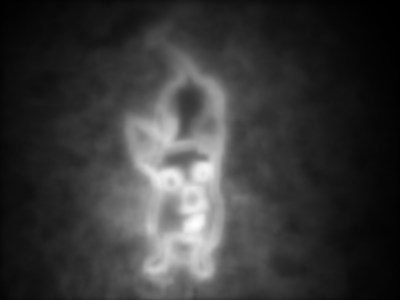}}%
\subfigure{\includegraphics[width=0.1\textwidth]{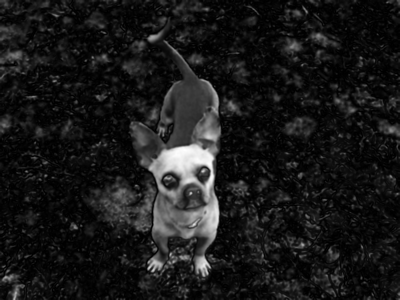}}%
\subfigure{\includegraphics[width=0.1\textwidth]{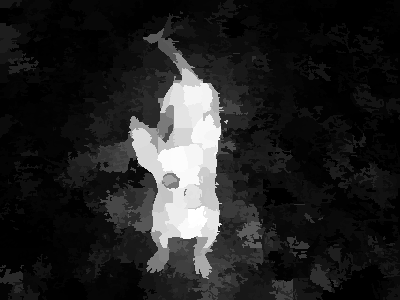}}%
\subfigure{\includegraphics[width=0.1\textwidth]{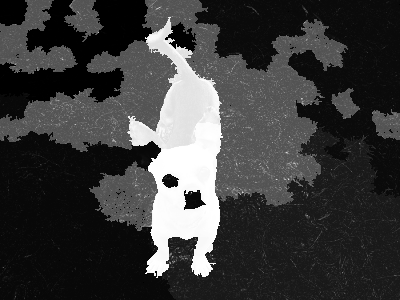}}%
\subfigure{\includegraphics[width=0.1\textwidth]{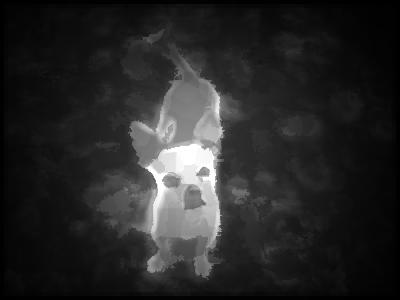}}%
\subfigure{\includegraphics[width=0.1\textwidth]{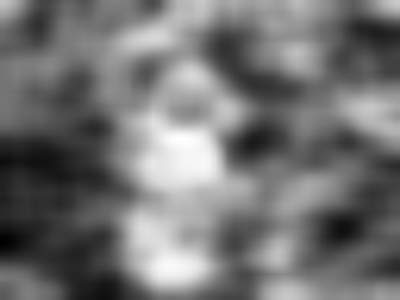}}%
\subfigure{\includegraphics[width=0.1\textwidth]{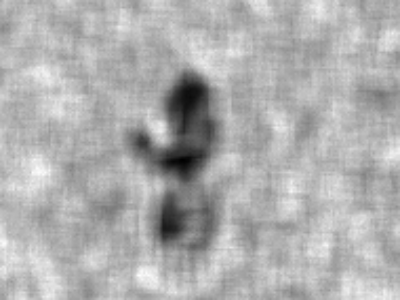}}%
\subfigure{\includegraphics[width=0.1\textwidth]{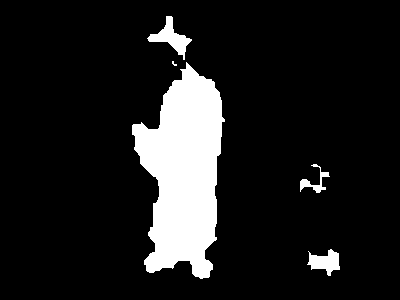}}\\
\subfigure{\includegraphics[width=0.1\textwidth]{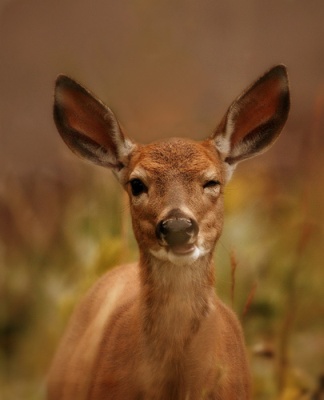}}%
\subfigure{\includegraphics[width=0.1\textwidth]{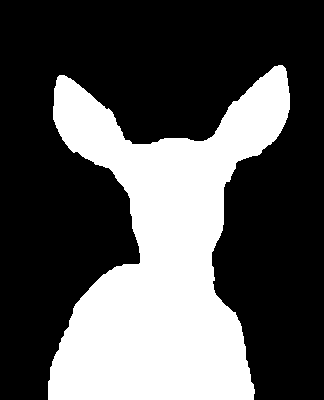}}%
\subfigure{\includegraphics[width=0.1\textwidth]{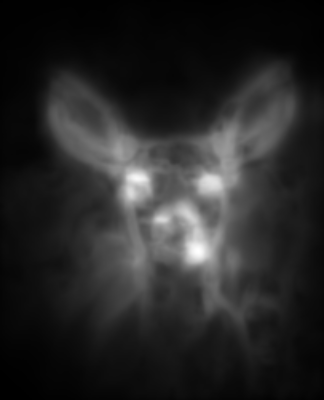}}%
\subfigure{\includegraphics[width=0.1\textwidth]{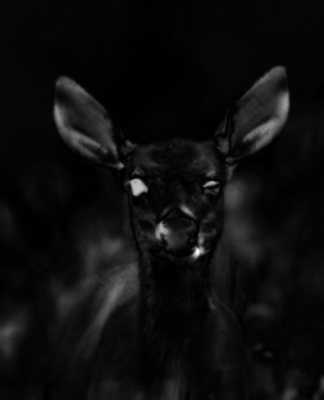}}%
\subfigure{\includegraphics[width=0.1\textwidth]{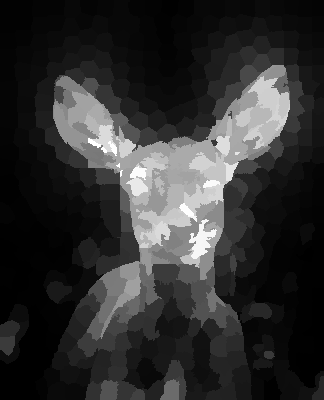}}%
\subfigure{\includegraphics[width=0.1\textwidth]{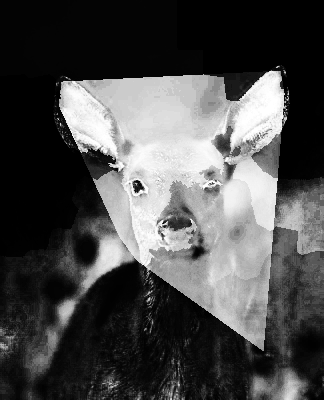}}%
\subfigure{\includegraphics[width=0.1\textwidth]{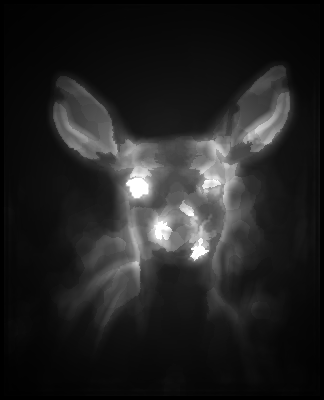}}%
\subfigure{\includegraphics[width=0.1\textwidth]{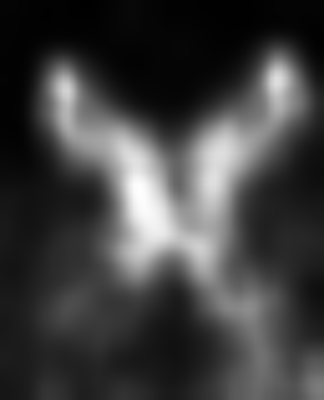}}%
\subfigure{\includegraphics[width=0.1\textwidth]{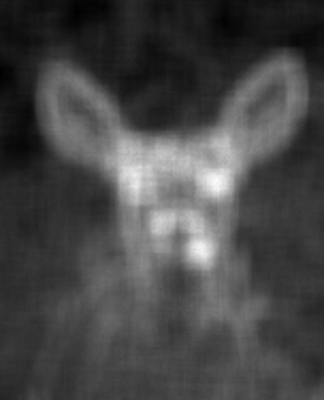}}%
\subfigure{\includegraphics[width=0.1\textwidth]{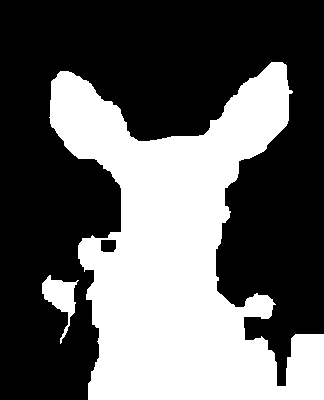}}\\
\subfigure{\includegraphics[width=0.1\textwidth]{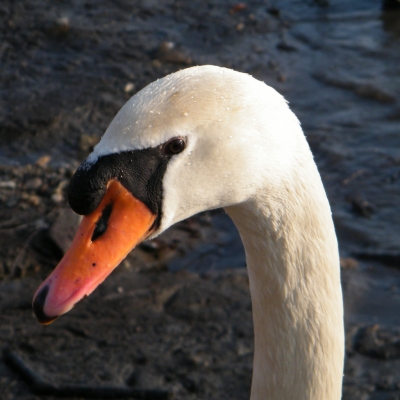}}%
\subfigure{\includegraphics[width=0.1\textwidth]{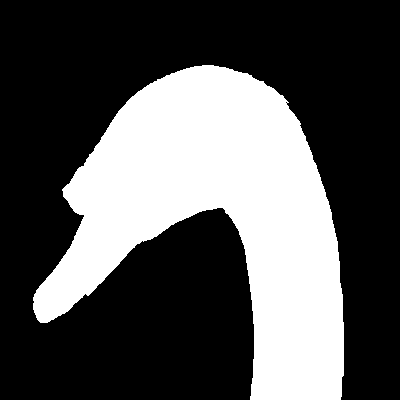}}%
\subfigure{\includegraphics[width=0.1\textwidth]{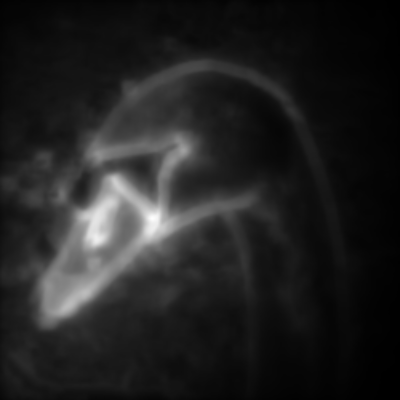}}%
\subfigure{\includegraphics[width=0.1\textwidth]{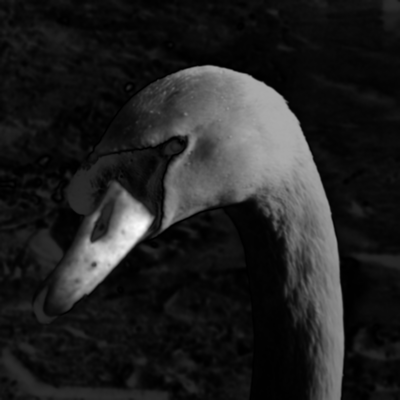}}%
\subfigure{\includegraphics[width=0.1\textwidth]{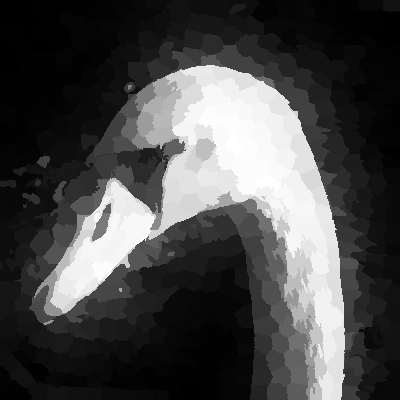}}%
\subfigure{\includegraphics[width=0.1\textwidth]{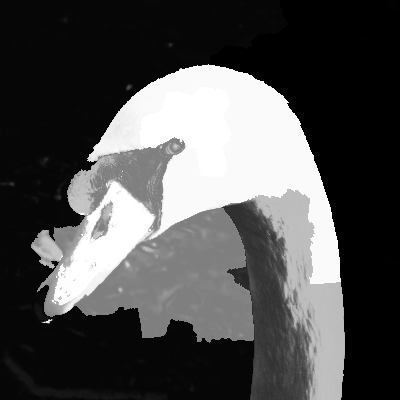}}%
\subfigure{\includegraphics[width=0.1\textwidth]{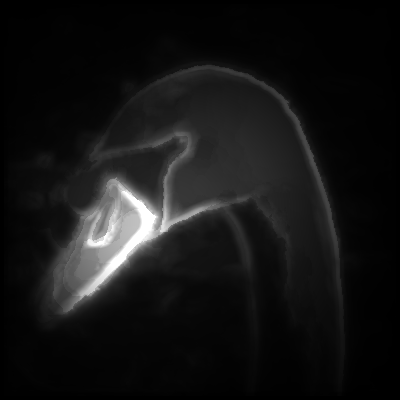}}%
\subfigure{\includegraphics[width=0.1\textwidth]{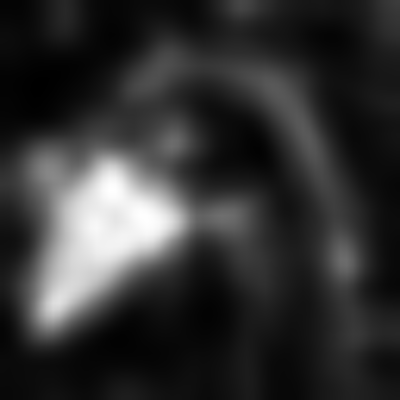}}%
\subfigure{\includegraphics[width=0.1\textwidth]{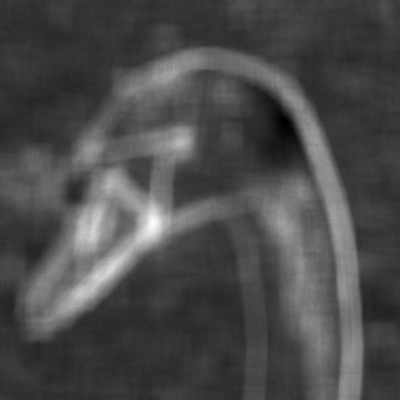}}%
\subfigure{\includegraphics[width=0.1\textwidth]{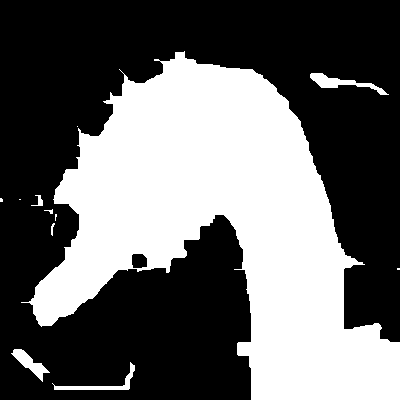}}\\
\subfigure{\includegraphics[width=0.1\textwidth]{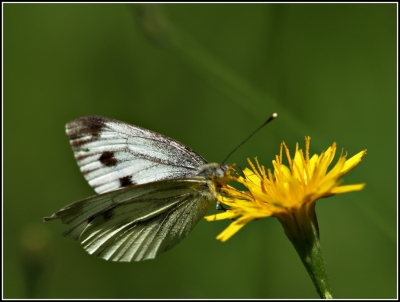}}%
\subfigure{\includegraphics[width=0.1\textwidth]{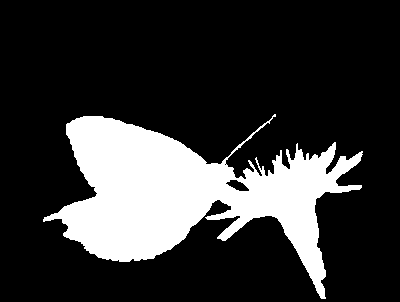}}%
\subfigure{\includegraphics[width=0.1\textwidth]{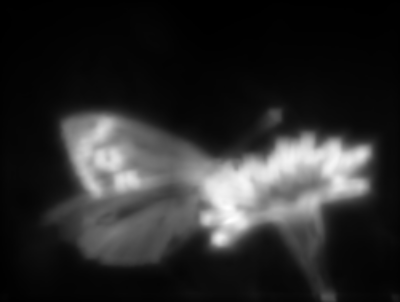}}%
\subfigure{\includegraphics[width=0.1\textwidth]{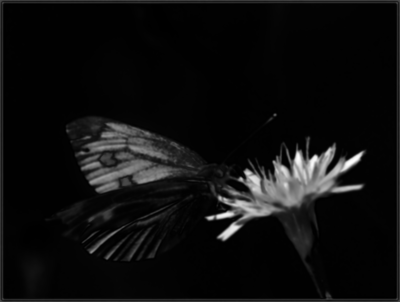}}%
\subfigure{\includegraphics[width=0.1\textwidth]{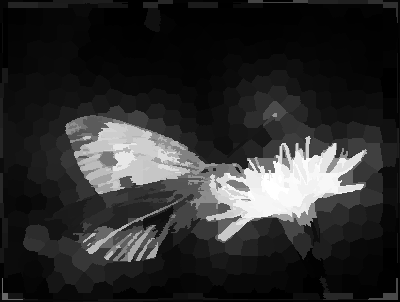}}%
\subfigure{\includegraphics[width=0.1\textwidth]{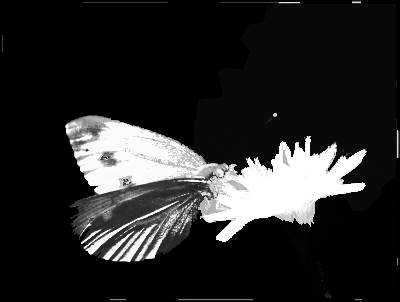}}%
\subfigure{\includegraphics[width=0.1\textwidth]{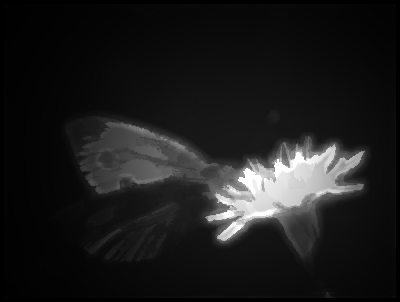}}%
\subfigure{\includegraphics[width=0.1\textwidth]{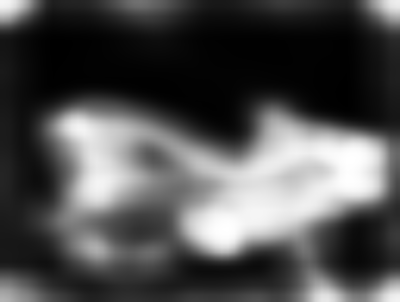}}%
\subfigure{\includegraphics[width=0.1\textwidth]{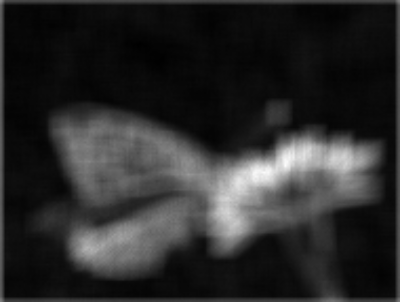}}%
\subfigure{\includegraphics[width=0.1\textwidth]{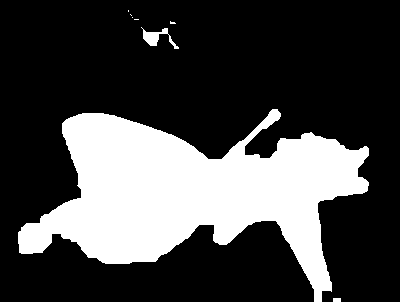}}\\
\subfigure{\includegraphics[width=0.1\textwidth]{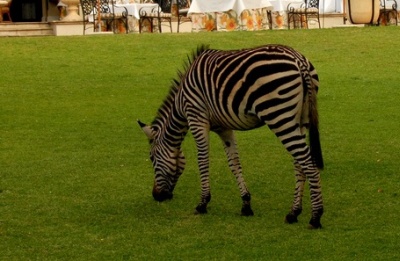}}%
\subfigure{\includegraphics[width=0.1\textwidth]{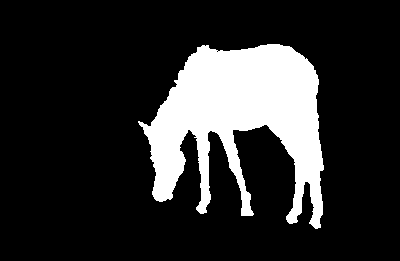}}%
\subfigure{\includegraphics[width=0.1\textwidth]{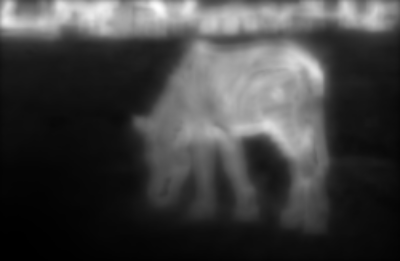}}%
\subfigure{\includegraphics[width=0.1\textwidth]{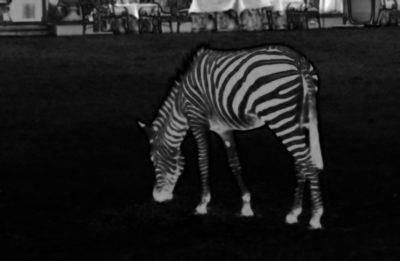}}%
\subfigure{\includegraphics[width=0.1\textwidth]{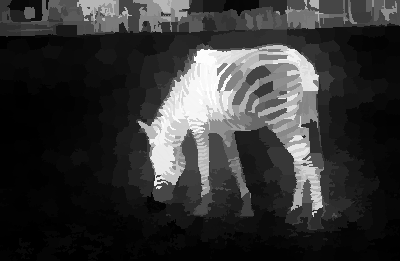}}%
\subfigure{\includegraphics[width=0.1\textwidth]{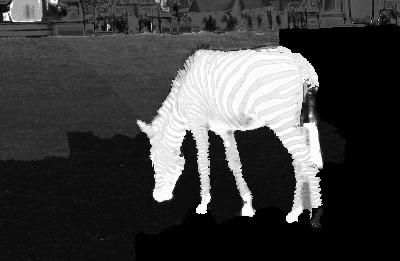}}%
\subfigure{\includegraphics[width=0.1\textwidth]{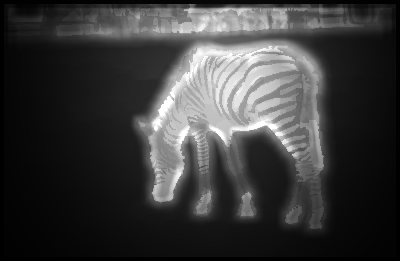}}%
\subfigure{\includegraphics[width=0.1\textwidth]{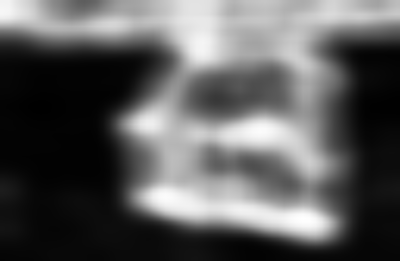}}%
\subfigure{\includegraphics[width=0.1\textwidth]{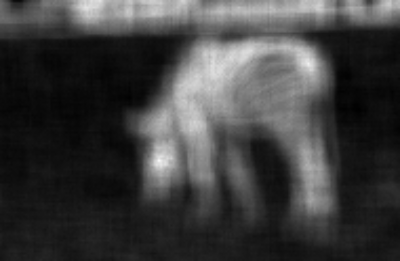}}%
\subfigure{\includegraphics[width=0.1\textwidth]{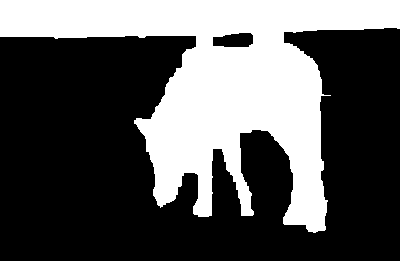}}\\
\subfigure{\includegraphics[width=0.1\textwidth]{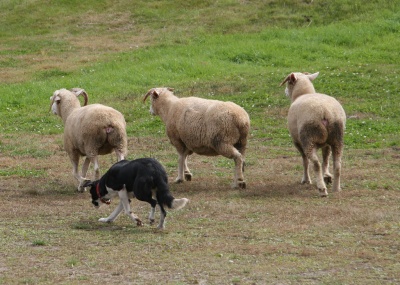}}%
\subfigure{\includegraphics[width=0.1\textwidth]{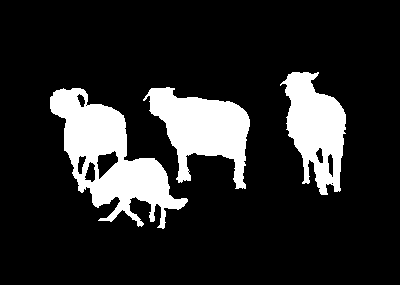}}%
\subfigure{\includegraphics[width=0.1\textwidth]{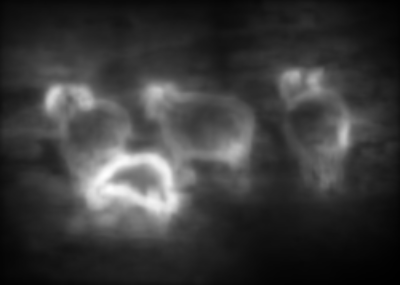}}%
\subfigure{\includegraphics[width=0.1\textwidth]{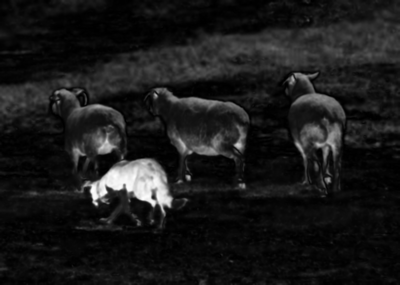}}%
\subfigure{\includegraphics[width=0.1\textwidth]{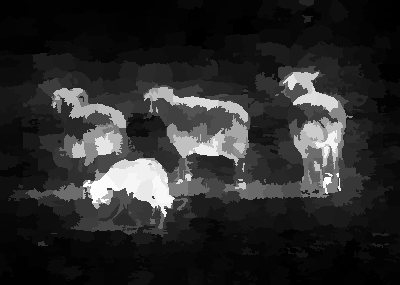}}%
\subfigure{\includegraphics[width=0.1\textwidth]{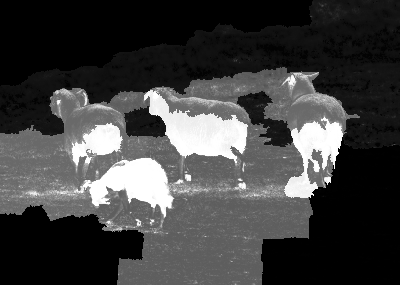}}%
\subfigure{\includegraphics[width=0.1\textwidth]{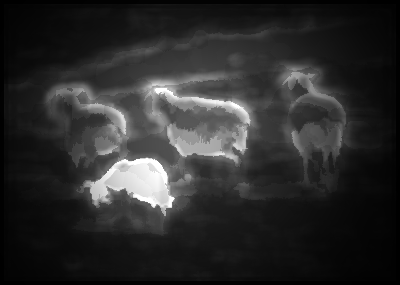}}%
\subfigure{\includegraphics[width=0.1\textwidth]{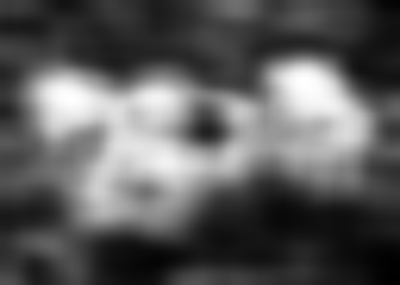}}%
\subfigure{\includegraphics[width=0.1\textwidth]{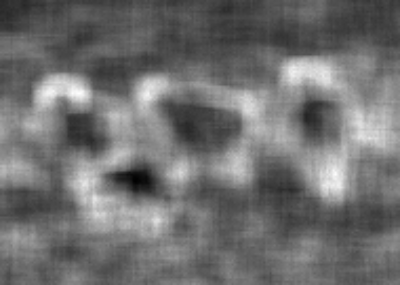}}%
\subfigure{\includegraphics[width=0.1\textwidth]{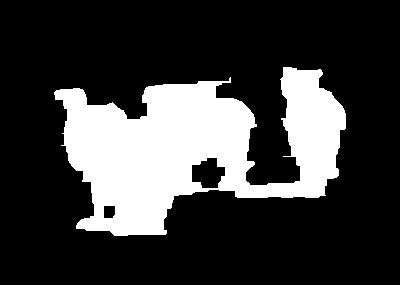}}\\
\subfigure{\includegraphics[width=0.1\textwidth]{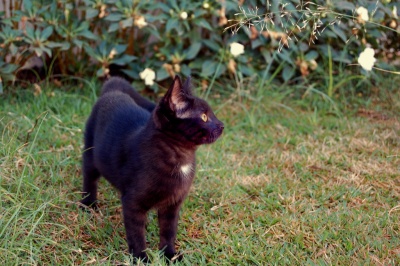}}%
\subfigure{\includegraphics[width=0.1\textwidth]{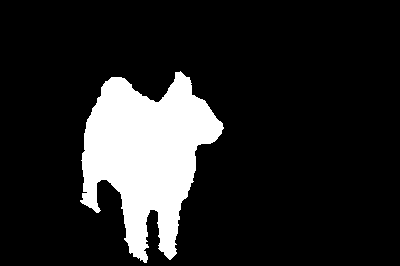}}%
\subfigure{\includegraphics[width=0.1\textwidth]{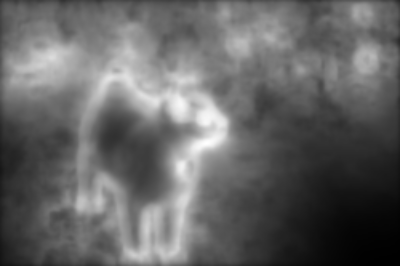}}%
\subfigure{\includegraphics[width=0.1\textwidth]{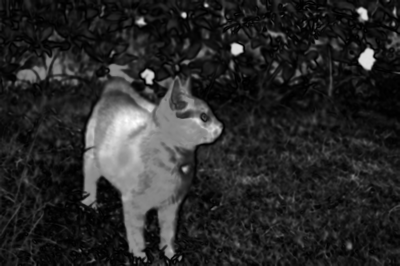}}%
\subfigure{\includegraphics[width=0.1\textwidth]{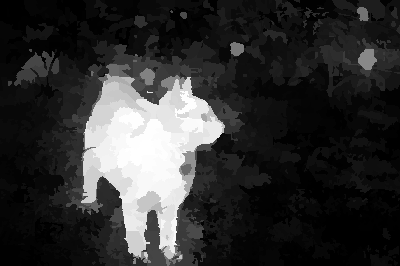}}%
\subfigure{\includegraphics[width=0.1\textwidth]{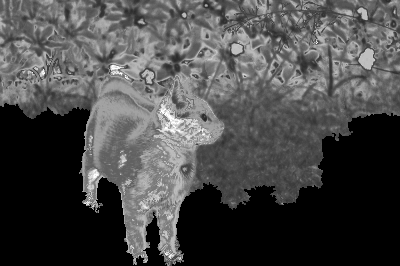}}%
\subfigure{\includegraphics[width=0.1\textwidth]{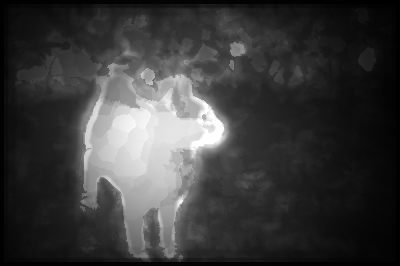}}%
\subfigure{\includegraphics[width=0.1\textwidth]{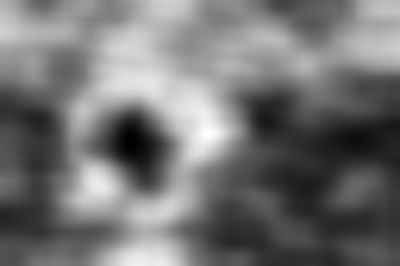}}%
\subfigure{\includegraphics[width=0.1\textwidth]{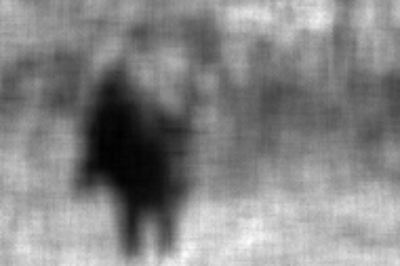}}%
\subfigure{\includegraphics[width=0.1\textwidth]{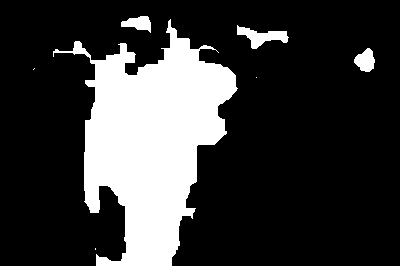}}\\%
\caption{Saliency Maps obtained form  various methods. Columns (a) - Original Image, (b) - Ground Truth, (c) - CA \cite{goferman2012context}, (d) - FT \cite{achanta2009frequency}, (e)- HDCT \cite{kim2014salient}, (f) - LMLC  \cite{xie2013bayesian}, (g) - PCA \cite{xie2013bayesian}, (h) SeR - \cite{seo2009static}, (f) SUN - \cite{zhang2008sun} and (j) OURS - Our Method.}
\label{Fig:4}
\end{figure*}

\section{Conclusions}

In this paper we discuss an approach to fuse the saliency maps obtained at different gaussian scales of an image. The reason that our method should work better than other methods are grounded on two factors. First, that we detected the saliency maps at different scale space. This corresponds to the working principle of human vision on which various descriptors like SIFT or SURF work. Second, we used dictionary learning to extract the non - linear features from image patches. This is faster to train and since we do it patchwise it ensures that local regions are merged non-uniformly according to the the surrounding patch's pixels' saliency scores. Thus, it should perform better than the previous methods. 

\section{Future Work}
Since we still need to optimize the hyperparameters and need to train the weights and dictionary properly for obtaining the good saliency maps. We are still very far from getting good results. Moreover, we also propose to test our results on the other datasets too, which we could not be able to do due to the lack of time. Also, instead of gaussian scale space, there is a possibility to use the scale space in which images are filtered using bilateral filter. This is because the bilateral filter smooths regions like gaussian filters but is better than gaussian filtering because it preserves the edges, thus there is a possibility it could provide the better saliency map.

\end{document}